\documentclass{article}
\usepackage{spconf,amsmath,graphicx}
\usepackage{threeparttable}

\usepackage{booktabs}
\usepackage{graphicx}
\usepackage{tabulary}
\usepackage{threeparttable}
\usepackage[numbers]{natbib}
\usepackage{hyperref}

\title{Improved ICH classification using task-dependent learning}
\name{Amir Bar$^1$, Michal Mauda Havakuk$^2$, Yoni Turner$^3$, Michal Safadi$^1$, Eldad Elnekave$^{1,4}$}
\address{$^1$Zebra Medical Vision Inc.\\ 
$^2$Tel-Aviv Medical Center, $^3$Shaare Zedek Medical Center, $^4$Rabin Medical Center}

\begin{document}

\maketitle
\begin{abstract}
Head CT is one of the most commonly performed imaging studied in the Emergency Department setting and Intracranial hemorrhage (ICH) is among the most critical and time-sensitive findings to be detected on Head CT.  We present BloodNet, a deep learning architecture designed for optimal triaging of Head CTs, with the goal of decreasing the time from CT acquisition to accurate ICH detection.  The BloodNet architecture incorporates dependency between the otherwise independent tasks of segmentation and classification, achieving improved classification results. AUCs of 0.9493 and 0.9566 are reported on held out positive-enriched and randomly sampled sets comprised of over 1400 studies acquired from over 10 different hospitals. These results are comparable to previously reported results with smaller number of tagged studies. 
\end{abstract}
\begin{keywords}
Deep Learning, Segmentation, ICH, Hemorrhage, Classification
\end{keywords}
\section{Introduction}
\label{sec:intro}

Intracranial hemorrhage (ICH) is a critical finding seen in various clinical circumstances spanning major trauma to spontaneous intracranial aneurysmal rupture. Early and accurate detection is essential in achieving optimal outcomes. An AI-facilitated first read of CT brains could provide value by detecting subtle bleeds which might go unrecognized, as well as providing triage-service to prioritize positively-flagged studies for expert radiologist review. 

In recent years, convolutional neural networks (CNN's) have been successfully designed to detect various pathologies in medical imaging \cite{textray, ccs, radbot, cf}. Previously reported deep-learning infrastructures for automatic ICH detection have based ICH prediction upon either the the entire 3D Head CT volume  \cite{deep3dconv} or each 2D CT slice \cite{qureai,radnet}. While the former potentially utilizes a larger amount of data, it is at the cost of relatively weak supervision due to the high dimensionality of the input volume. The second approach requires a substantial tagging effort due to tedious annotation of every relevant slice in the scan. 

Jnawali et al~\cite{deep3dconv} assembled a dataset of 40k studies and preprocessed it to a fixed input size. It was then used for the training of a 3D convolution~\cite{3dconvaction} classification pipeline and reported to have an AUC of 0.86 using a single model. Additional work was in~\cite{qureai}, in which the authors utilized a large dataset of 6k studies tagged slice-wise by radiologists for training. To localize the findings, the authors had to annotate the slices pixel-wise to create the masks necessary in order to train a UNet \cite{unet} architecture for segmentation. They report AUC of 0.9419 for the classification part. In \cite{radnet}, the authors used multiple segmentation auxiliary losses to leverage the pixel-wise information and aggregated the 3D volumetric decision using LSTM \cite{hochreiter1997long}. 

The present report describes integration of both classification and segmentation of an image in a single network, utilizing the pixel-wise prediction to improve the 3D volumetric ICH classification result. BloodNet is a CNN architecture which explicitly incorporates the pixel-wise prediction through modeling the dependency between the classification and segmentation task. 
\begin{figure}[htb]
\begin{minipage}[b]{1.0\linewidth}
  \centering
  \centerline{\includegraphics[width=1\textwidth]{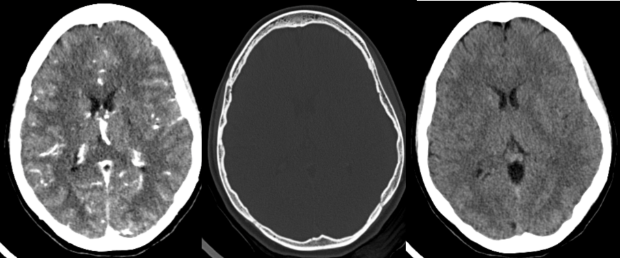}}
  \caption{Head CT typical scans. Left to right: Head CT with contrast, Head CT with bone enhancing reconstruction and non contrast Head CT. The proposed system operates on the last type.}
  \medskip
  \label{fig:HeadCT}
\end{minipage}
\end{figure}

\begin{figure}[htb]
\begin{minipage}[b]{1.0\linewidth}
  \centering
  \centerline{\includegraphics[width=1\textwidth]{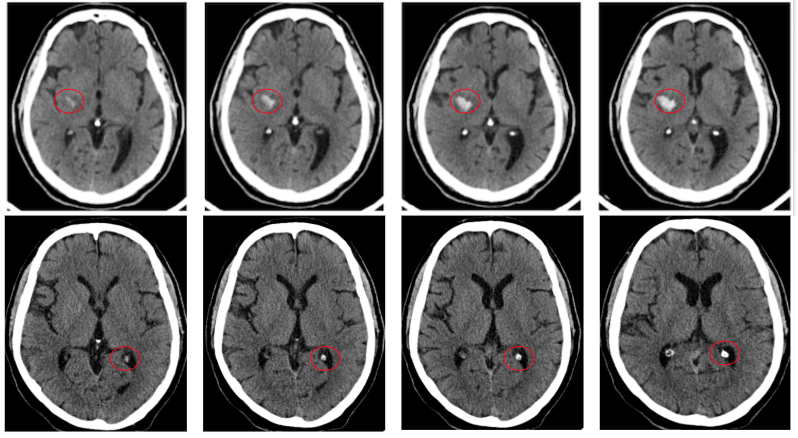}}
  \caption{Context is necessary with respect to ICH. Top row: Parenchymal hemorrhage. Bottom row: calcification.}
  \medskip
  \label{fig:Context}
\end{minipage}
\end{figure}

\begin{table}
    \setlength{\tabcolsep}{18pt}
    \begin{center}
    \begin{threeparttable}
    \caption{Data}\label{tab1}
        \begin{tabular}{lcccc}
            \toprule
            \toprule
            & Train & Validation\\
            & (1)  & (2)\\
            \midrule
            \quad Positive & 3953 & 1815 & \\
            \quad Negative & 22122 	& 4141 & \\
            \bottomrule
        \end{tabular}
    \begin{tablenotes}[para,flushleft]
        Every cell represents the number of tagged slices. All slices were manually pixel wise annotated for positive ICH on 175 ICH positive and 102 on ICH negative scans.
    \end{tablenotes}
    \end{threeparttable}
    \end{center}
\end{table}

\section{Materials and Methods}

For training and validation, 175 non-contrast CT brain studies with ICH-positive radiology reports were reviewed by at least one expert radiologist who validated the existence of the reported ICH and manually segmented it.  An ICH-negative dataset including 102 CTs was also assembled. For validation we use only positive studies, which contain both positive and negative slices. Testing was performed on two datasets totaling 1,426 expert-validated studies, including an enriched (67\% ICH positive) and randomly sampled (16\% positive) set. Every study was tagged by a single expert radiologist while multiple experts participated in the tagging.

\begin{figure}[htb]
\begin{minipage}[b]{1 \linewidth}
  \centering
    \centerline{\includegraphics[width=1\textwidth]{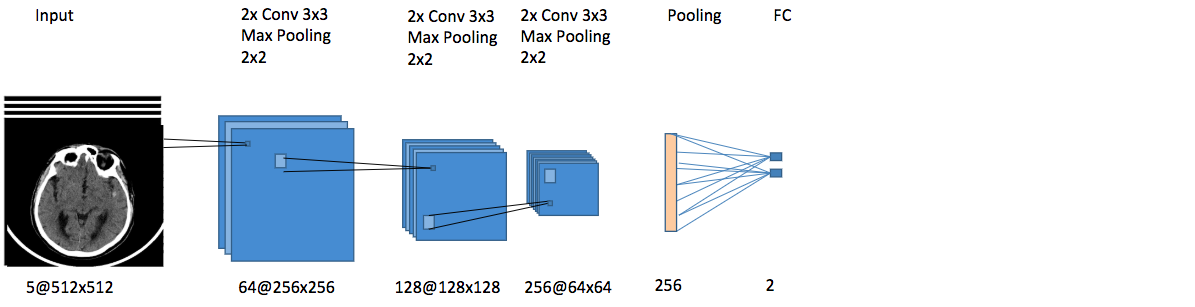}}
    \caption{Single task, classification. Predicting whether an input has positive indication for ICH.}
  \label{fig:Classification}
\end{minipage}
\end{figure}

\begin{figure}[htb]
\begin{minipage}[b]{\linewidth}
  \centering
  \centerline{\includegraphics[width=1.\textwidth]{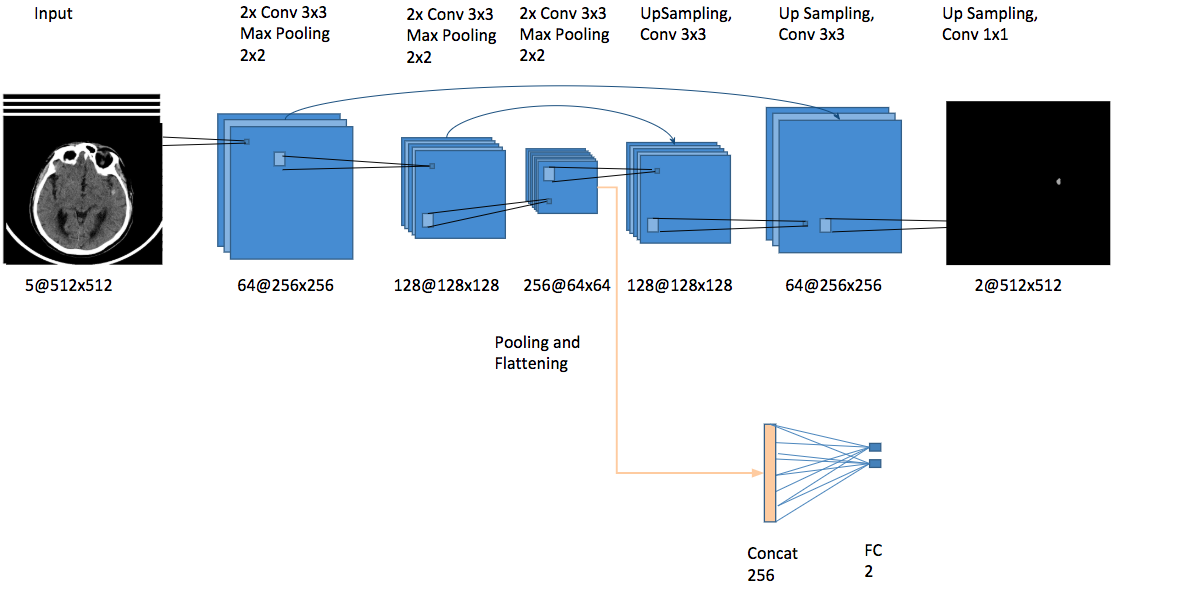}}
  \caption{Multi-task, classification and segmentation. The shared encoder learns a representation suitable for both tasks.}
    \label{fig:SegmentationClassification}
\end{minipage}
\end{figure}

\begin{figure*}[htb]
\begin{minipage}[b]{\linewidth}
  \centering
    \centerline{\includegraphics[width=1\textwidth]{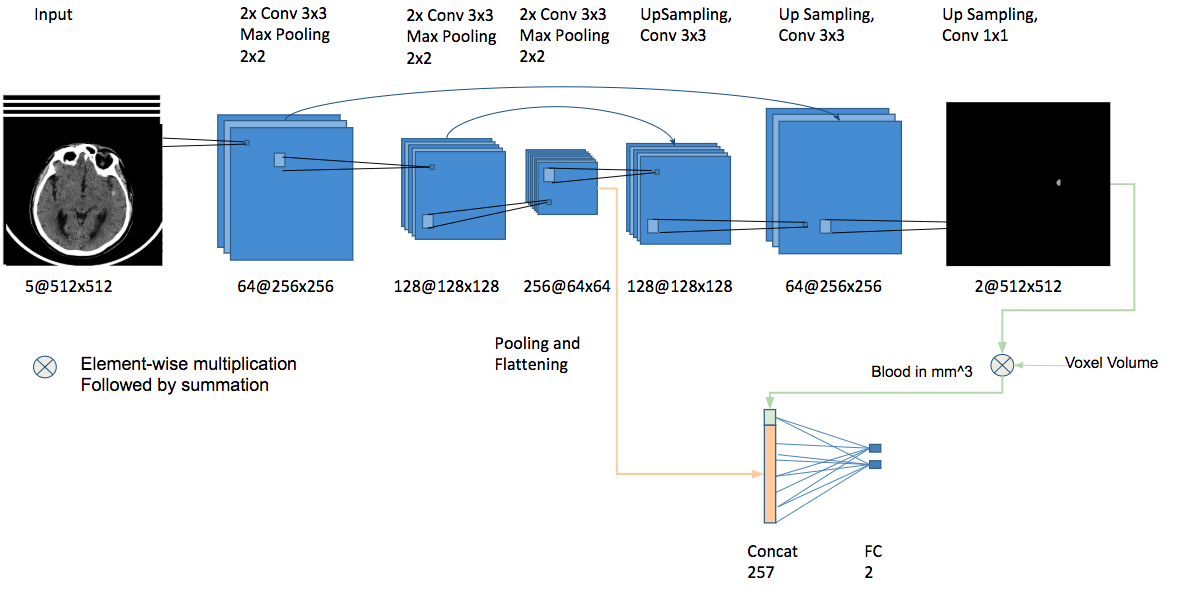}}
  \caption{Task dependent, classification is dependent on segmentation. The segmentation branch result probabilities are multiplied by the voxel volume of the input, summed and finally concatenated to the bottleneck as an additional feature for classification.}
 \label{fig:ClassificationDependentSegmentation}
 \end{minipage}
\end{figure*}
The present report describes a new pipeline for CT-based ICH classification intended for enhanced triage. The setup relies on the learning of both classification and segmentation, having demonstrated that the segmentation task provides synergistic support to the ICH classification task. A high level description of our architecture is described in Figure ~\ref{fig:ClassificationDependentSegmentation}.

To exploit the volumetric nature of ICH, the input number of slices was set as 5 consecutive axial CT slices, allowing for better detection of true ICH. We empirically observed that the learned models better distinguish artifacts and hemorrhages, which may look similar on a single slice but commonly appear differently over consecutive slices. We show example for the advantage of such context in Fig ~\ref{fig:Context}. Additional preprocessing included the utilization of standard brain-windowing. Since we empirically observed that a hemorrhage might be very small, we kept the input slices in the full 512x512 CT resolution. 

Given the input slices we first base our approach on performing classification alone, using the architecture in Figure \ref{fig:Classification}. Hence our classification loss is:

$$L_{classification} = \frac{1}{m}\Sigma^m_{i=1} CE(y_i, \hat{y_i})$$

Where $y_i$ is the ground truth label, $\hat{y_i}$ is the prediction of the $i$-th sample, $m$ is the number of samples and $CE$ is the binary cross entropy function:

$$CE(y, \hat{y}) = y log\hat{y} + (1-y) \cdot log(1-\hat{y})$$

Considering the clear advantages of multi-task learning reported in recent research \cite{he2017mask,radnet}, we modified the architecture and added a decoder to enable the multi-task learning scenario of classification and segmentation (see Figure \ref{fig:SegmentationClassification}). We also added an auxiliary segmentation loss:

$$L_{segmentation} = \frac{1}{m\cdot h\cdot w}\Sigma^m_{i=1}\Sigma^{h}_{j=1}\Sigma^{w}_{k=1} CE(y_{ijk}, \hat{y_{ijk}})$$

Where $h$ and $w$ are the height and width of input slice, $y_{ijk}$ is the pixel in the spatial position $j,k$ of the $i$th sample.

Our final loss is thus:
$$L = (1-\lambda)L_{classification} +\lambda \cdot L_{segmentation}$$

Finally, instead of implicitly using the segmentation information as supervision, we explicitly design the architecture to utilize the segmentation information to support classification. More specifically, we sum over the decoder network segmentation prediction, multiply by the voxel volume and concatenate the approximation of blood in $mm^3$ as a feature in the classification branch. 

To train this architecture, we employ three steps. First, we train the segmentation branch alone. Then, we freeze all weights and train only the last fully connected layer of the classification branch. Finally, we train the entire architecture for both classification and segmentation in an end-to-end manner.  Respectively, we use $\lambda=1$, $\lambda=0$, $\lambda=0.5$ in the loss equation. In all our experiments we use the Adam optimizer with learning rate of $1e-4$ and exponential decay of $0.96$. All architectures were implemented in Tensorflow and trained using 4 Nvidia Tesla K80 GPUs. In inference, given a study, we compute the probability for ICH over every slice and use the maximal probability as the study probability for ICH.

\section{Results}

We choose the best architecture using AUC over validation set. Table \ref{table:validation} provides comparison between models. We then evaluated on two different held out test sets, a positive enriched and a randomly sampled sets. The advantage of a positive enriched set is in representation of different types of ICHs as well as ICHs which are less prevalent. To collect this set we used a textual search over radiology reports. Since such data collection method might present a bias towards a specific search criteria, we also collected a randomly sampled set. We assume that in the randomly sampled set the cases in radiologists daily routine are well represented. We report AUCs of 0.9493 and 0.9566 over the enriched and randomly sampled tests set. Table ~\ref{tab3} provides further information. A manual review of false positives showed propensity to aberrantly misclassify calcified hemangiomas, dystrophic parenchymal calcifications and basal ganglial calcifications.

\begin{table}
\begin{minipage}[b]{\linewidth}

    \begin{threeparttable}
        \caption{Test sets results}\label{tab3}
        \begin{tabular}{lcccc}
            \toprule
            \toprule
            & \#Studies & \%ICH & AUC \\
            & (1)  & (2) & (3)\\
            \midrule
            \quad Test-Enriched & 608 	&	67\%	&	0.9493&  \\
            \quad Test-Random & 818 	&	16\%	&	0.9566&  \\
            \bottomrule
        \end{tabular}
        \begin{tablenotes}[para,flushleft]
            Every cell represents the number of tagged studies. These studies were tagged only on study level. These studies were held out during training and validation with respect to patient.
        \end{tablenotes}
    \end{threeparttable}
\end{minipage}
\end{table}

\begin{table*}
  \centering

    \begin{center}
    \begin{threeparttable}
        \caption{Comparison between models}\label{tab1}
        \begin{tabular}{lcc}
            \toprule
            \toprule
            Network & AUC & 0.95 CI\\
            & (1) & (2) \\
            \midrule
            i. Baseline \\
            \quad ResNet50 \cite{he2016deep} & 0.9159 & [0.9081, 0.9236]\\
            \\
            ii. BloodNet \\
            \quad Single task, classification  & 0.9453 & [0.9395, 0.9512] \\
            \quad Multi task, classification and segmentation & 0.9411 & [0.9352, 0.9471] \\
            \quad Task dependent, segmentation dependent classification & \bf 0.9658 & [0.9611, 0.9704]\\
            
            \bottomrule
            \label{table:validation}
        \end{tabular}
        \begin{tablenotes}[para,flushleft]
        Ablation studies of results over validation set slices. Mean AUC and CI are are computed using bootstrap $(n=10^4)$.
        \end{tablenotes}
    \end{threeparttable}
    \end{center}
\end{table*}
\begin{figure*}[htb]

\begin{minipage}[b]{1.0\linewidth}
  \centering
  \centerline{\includegraphics[width=10cm]{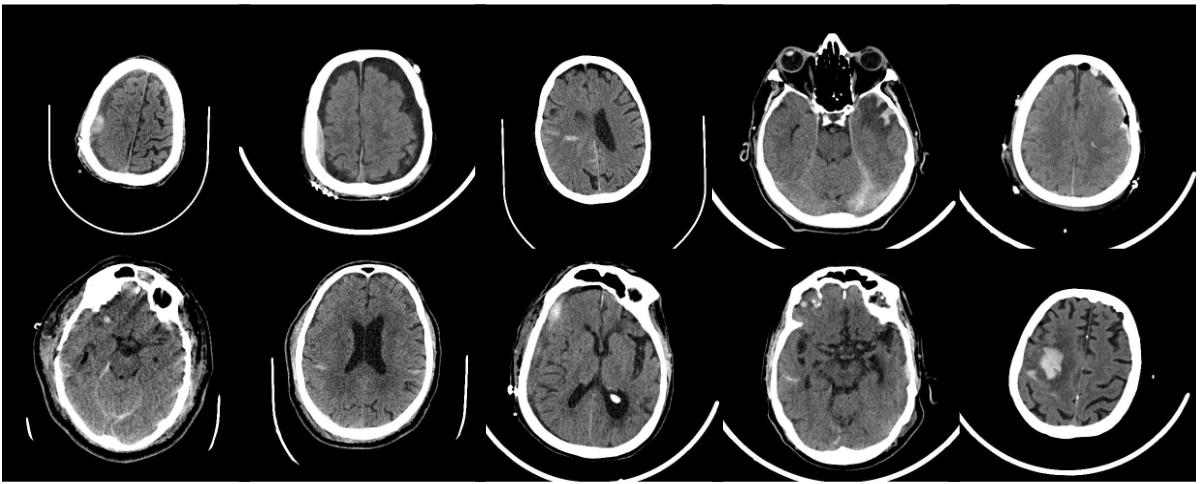}}
  
  \medskip
  \caption{Visualizations of studies which received high probability for ICH on validation set.}
\end{minipage}

\end{figure*}
\begin{figure}[htb]
\begin{minipage}[b]{1\linewidth}
  \centering
  \centerline{\includegraphics[width=8.0cm]{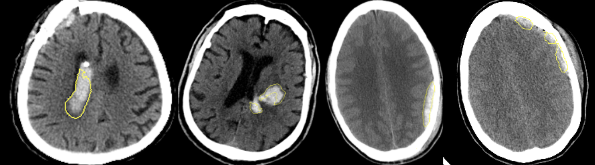}}
  \caption{Selected Segmentation results example.}
\end{minipage}
\end{figure}

\section{Discussion}

This work provides further evidence to support the approach of utilizing pixel wise annotated data for classification. However, our results indicate that relying on the multi-task setting alone might not be enough to yield a significant improvement in performance for classification. In BloodNet, we explicitly model a segmentation dependent classification, resulting in design that fully leverages the dense pixel wise supervision to boost classification performance. It has the advantage of both classification and localization of the acute finding and while classification is most important in a triage system, the localization provides reasoning hence crucial for a radiologist to have a better understanding of the prediction.

\newpage
\begin{figure}
\begin{minipage}[b]{1\linewidth}
  \centering
  \centerline{\includegraphics[width=7.0cm]{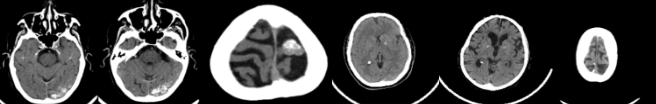}}
  \caption{Visualizations of negative studies which received high probability. Mengiomas and calcifications.}
\end{minipage}
\end{figure}

\section*{Acknowledgement}
The authors would like to thank Orna Bregman, Assaf Pinhasi, Jonathan Laserson, David Chettrit, Chen Brestel, Eli Goz, Phil Teare, Tomer Meir,  Rachel Wities, Amit Oved, Raouf Muhamedrahimov and Eyal Toledano for helpful comments and discussions during this research.

\clearpage
\bibliographystyle{IEEEbib}
\bibliography{biblio}

\end{document}